\documentclass{article}

\usepackage{arxiv}

\usepackage[utf8]{inputenc}
\usepackage[T1]{fontenc}
\usepackage[english]{babel}
\usepackage[numbers]{natbib}
\usepackage{amsmath,amssymb,amsthm}
\usepackage{graphicx}
\usepackage{booktabs}
\usepackage{multirow}
\usepackage{tabularx}
\usepackage{textcomp}
\usepackage{xcolor}
\usepackage{hyperref}
\usepackage{url}
\usepackage{microtype}
\usepackage{doi}
\usepackage[labelfont=bf]{caption}
\usepackage{titlesec}
\titleformat{\subsection}[hang]{\normalfont\large}{\thesubsection}{1em}{}
\titleformat{\subsubsection}[hang]{\normalfont\normalsize\itshape}{\thesubsubsection}{1em}{}

\graphicspath{{figs/}}

\providecommand{\tblwidth}{\linewidth}

\newtheorem{definition}{Definition}

\newtheorem*{axiomstar}{Axiom}

\renewcommand{\shorttitle}{TOTEN: structure-preserving representation of physical quantities and technical notation in PT-BR}

\hypersetup{
  pdftitle={TOTEN: A Knowledge-Based System for Structure-Preserving Representation of Physical Quantities and Technical Notation in Brazilian Portuguese},
  pdfsubject={cs.CL, cs.AI},
  pdfauthor={Antonio de Sousa Leit\~ao Filho, Allan Kardec Duailibe Barros Filho, Fabr\'icio Saul Lima, Selby Mykael Lima dos Santos, Rejani Bandeira Vieira Sousa},
  pdfkeywords={Knowledge-based systems, Structure-preserving representation, Physical quantity extraction, Ontological tokenization, Quantitative reasoning, Brazilian Portuguese NLP},
}

\title{TOTEN: A Knowledge-Based System for Structure-Preserving Representation of Physical Quantities and Technical Notation in Brazilian Portuguese}

\author{
  \href{https://orcid.org/0009-0002-1705-3611}{\includegraphics[scale=0.06]{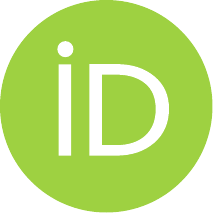}\hspace{1mm}Antonio de Sousa Leit\~ao Filho}$^{1,2,*}$
  \\[0.15em]
  \href{https://orcid.org/0000-0002-1654-0955}{\includegraphics[scale=0.06]{orcid.pdf}\hspace{1mm}Allan Kardec Duailibe Barros Filho}$^{2}$
  \\[0.15em]
  \href{https://orcid.org/0009-0005-1837-8751}{\includegraphics[scale=0.06]{orcid.pdf}\hspace{1mm}Fabr\'icio Saul Lima}$^{1,2}$
  \\[0.15em]
  \href{https://orcid.org/0009-0006-6627-6503}{\includegraphics[scale=0.06]{orcid.pdf}\hspace{1mm}Selby Mykael Lima dos Santos}$^{1,2}$
  \\[0.15em]
  \href{https://orcid.org/0009-0000-7888-7324}{\includegraphics[scale=0.06]{orcid.pdf}\hspace{1mm}Rejani Bandeira Vieira Sousa}$^{1,3}$
  \\[0.5em]
  \small $^{1}$Aia Context, S\~ao Lu\'is, Brazil
  \\[0.1em]
  \small $^{2}$Universidade Federal do Maranh\~ao, S\~ao Lu\'is, Maranh\~ao, Brazil
  \\[0.1em]
  \small $^{3}$Universidade de S\~ao Paulo, S\~ao Paulo, Brazil
  \\[0.1em]
  \small $^{*}$Corresponding author:
  \href{mailto:antonio@aiacontext.com}{\texttt{antonio@aiacontext.com}}
}

\begin{document}

\maketitle

\begin{abstract}
Artificial intelligence pipelines performing quantitative reasoning over technical text depend on input in which physical quantities, numbers, units, and symbolic expressions arrive intact; when these entities are fragmented at tokenization, the error propagates throughout the downstream system. Statistical tokenization by \textit{Byte-Pair Encoding}, optimized for vocabulary compression, is semantically blind to these entities and fragments them into lexically arbitrary subwords --- a problem aggravated in technical Brazilian Portuguese. We present \textsc{TOTEN}, a knowledge-based system that produces an input representation in which each technical entity is preserved as a whole, typed unit: rather than deriving vocabulary statistically, it classifies textual regions declaratively, governed by a formal ontology of engineering entities (OEE) that constitutes the engine of the system. The ontological core is formalized as the triple $\langle \mathcal{O}, \mathrm{classify}, \{\mathrm{inst}_\tau\}\rangle$ --- types, structural principles, composition relations, and preservable invariants; a classification function that maps raw text into typed regions; and an indexed family of instantiators that produces a self-descriptive structured representation. Integrity rests on deterministic coupling to three consolidated external authorities: \textit{Pint} (dimensional), the \textit{Unicode Character Database} (typographic), and \textit{RSLP} (Portuguese morphology). We evaluate the system on four properties verifiable by construction --- ontological atomicity, dimensional equivalence, typographic robustness, and numerical reconstruction --- over an internally generated and physically validated benchmark (\textit{EngQuant}, $N=800$) and four external corpora in Brazilian Portuguese ($N=1\,771$ cases eligible for numerical reconstruction), additionally reporting detection \emph{recall}. Compared to eight representative state-of-the-art systems, \textsc{TOTEN} achieves unit ontological atomicity in all contrasts and numerical reconstruction of $0.775$ to $0.904$ on external corpora, against $0.627$--$0.703$ for the best baseline (Quantulum3); on the internal benchmark, $0.780$ against $0.340$. Differences in atomicity and reconstruction are statistically significant (McNemar with Holm correction). The Spearman rank correlation between internal and external corpus rankings confirms the concurrent validity of the control benchmark. In dimensional equivalence, the system shows statistical parity with \textit{Pint}, the oracle from which it inherits dimensional authority. These results characterize \textsc{TOTEN} as a structurally faithful, auditable, and computationally inexpensive input layer for intelligent systems operating on technical knowledge, without dependency on generative models.
\end{abstract}

\keywords{Knowledge-based systems \and Structure-preserving representation \and Physical quantity extraction \and Ontological tokenization \and Quantitative reasoning \and Brazilian Portuguese NLP}

\section{Introduction}
\label{sec:introducao}

The symbolic representation of technical entities in scientific text remains an unsolved problem in contemporary language models. Statistical tokenization algorithms such as \textit{Byte-Pair Encoding} \citep{Sennrich2016}, \textit{WordPiece}, and \textit{SentencePiece} \citep{Kudo2018} are derived from predominantly English generalist corpora and produce vocabularies whose granularity is optimized for statistical compression, not semantic preservation. When applied to technical text in Brazilian Portuguese, these tokenizers fragment semantically atomic entities --- physical quantities, locale-specific numbers, compound dimensional units, symbolic expressions --- into subword sequences whose recomposition depends entirely on a downstream model a posteriori. Adjacent cases such as normative identifiers (NBR, ABNT) and hierarchical references to legal articles and paragraphs suffer from the same structural problem; their evaluation on an annotated open corpus is, however, left as an extension of this work.

Recent studies empirically document the consequences of this fragmentation. \citet{Singh2024} demonstrate that right-to-left digit grouping substantially increases arithmetic accuracy in GPT-3.5, indicating that digit structure preserved in the input is associated with better downstream arithmetic performance independently of parameter scale. \citet{Yang2025} catalogue systematic gaps in numerical reasoning whose origin is attributed to inadequate tokenization, not training. These findings motivate the design of an input representation that preserves the semantic structure of numbers and units; direct verification of the downstream effect on consumer models is outside the scope of this study and is left as future work.

Domain-specific literature on quantitative extraction for English \citep{Almasian2023,Zaratiana2024} offers partial solutions that do not adequately model the technical vocabulary of Brazilian Portuguese; dimensional libraries such as \textit{Pint} \citep{Grecco2022} and \textit{udunits-2} \citep{Hankin2020} operate on already-isolated unit strings, without performing textual recognition; generic entity recognition models \citep{Honnibal2020} are trained on categories such as person, location, and organization, ignoring technical-scientific vocabulary.

This work proposes an alternative grounded in \textbf{ontological engineering}. Rather than deriving vocabulary statistically, we explicitly declare a formal ontology of engineering entities (OEE), comprising primary types, structural principles, composition relations, and preservable invariants. Over this ontology, we define \textsc{TOTEN}\footnote{TOTEN --- \emph{Typed Ontological Tokenization}.}, a knowledge-based tokenization system operating in three functional layers and coupled to three consolidated external oracles. The scientific contributions of this work are:

\textbf{(C1)} A \emph{formalization} of ontological tokenization as the triple $\langle \mathcal{O}, \mathrm{classify}, \{\mathrm{inst}_\tau\}\rangle$, implementation-independent and amenable to evaluation via verifiable properties, categorically distinguishing it from statistical subword tokenization.

\textbf{(C2)} A \emph{formal ontology of engineering entities} (OEE) with primary types defined by intrinsic properties, \textbf{eight} structural principles expressed as axioms (Appendix~\ref{app:axiomas}) and declared composition relations, extensible under the \emph{open-for-extension, closed-for-modification} principle \citep{Meyer1997}.

\textbf{(C3)} A computationally inexpensive \emph{intrinsic evaluation} based on four properties verifiable by construction --- reporting, beyond detection (\emph{recall}), the four properties: atomicity, dimensional equivalence, typographic robustness, and numerical reconstruction --- replicated over five corpora (one physically validated internal and four external PT-BR), with oracle ablation and cross-corpus ranking consistency validation, demonstrating statistically significant advantage in atomicity and numerical reconstruction against eight state-of-the-art systems.

Section~\ref{sec:fundamentos} establishes the theoretical foundations in ontological engineering and tokenization. Section~\ref{sec:oee} formalizes the OEE. Section~\ref{sec:arquitetura} presents the architecture of \textsc{TOTEN}. Section~\ref{sec:representacao} characterizes the output language. Section~\ref{sec:avaliacao} describes the experimental protocol. Section~\ref{sec:resultados} presents the results. Section~\ref{sec:discussao} discusses implications. Section~\ref{sec:conclusao} concludes.

\section{Theoretical Foundations}
\label{sec:fundamentos}

\subsection{Ontological Engineering}

A formal ontology, in the sense established by \citet{Gruber1993}, is an explicit specification of a shared conceptualization of a domain. \citet{Studer1998} characterize ontological engineering as a discipline that produces reusable formal artifacts for knowledge representation, distinguishing \emph{lightweight} ontologies (taxonomies with few constraints) from \emph{heavyweight} ones (axiomatic, with vocabulary rigorously constrained by logical axioms). \citet{Guarino1998} introduces the criterion of \emph{ontological commitment} as a theory's obligation to the structure of the reality it describes.

We adopt the classical formalization of an ontology as a quadruple
\begin{equation}
\mathcal{O} = \langle \mathcal{T},\ \mathcal{P},\ \mathcal{R},\ \mathcal{I}\rangle,
\label{eq:ontologia}
\end{equation}
where $\mathcal{T}$ is the finite set of primary types, $\mathcal{P}$ is the set of structural principles (axioms), $\mathcal{R} \subseteq \mathcal{T} \times \mathcal{T}$ is the composition relation between types, and $\mathcal{I}$ is the set of invariants that any valid representation of instances must preserve. This formulation is compatible with \citep{Studer1998} and admits incremental extension: given $\mathcal{O}_n$ at version $n$, version $n+1$ satisfies $\mathcal{T}_n \subseteq \mathcal{T}_{n+1}$ and $\mathcal{P}_n \subseteq \mathcal{P}_{n+1}$, without any types or prior principles being removed.

\subsection{Tokenization: Formal Definition}

Let $\Sigma$ be a finite alphabet and $\Sigma^*$ the set of all finite strings over $\Sigma$. A \emph{tokenization} is a function
\begin{equation}
\mathrm{tok}: \Sigma^* \longrightarrow V^*,
\end{equation}
where $V$ is a token vocabulary. Two families are categorically distinct: \textbf{statistical tokenization}, in which $V$ is induced from a corpus $\mathcal{C}$ by a compression procedure (BPE, WordPiece, SentencePiece); and \textbf{ontological tokenization}, in which $V$ is a language $\mathcal{M}$ defined over an ontology $\mathcal{O}$, and the function factors into two components: $\mathrm{tok} = \mathrm{ext} \circ \mathrm{classify}$, where $\mathrm{classify}: \Sigma^* \to \mathcal{P}(\mathcal{R})$ identifies typed regions and $\mathrm{ext}: \mathcal{P}(\mathcal{R}) \to \mathcal{M}^*$ produces the structured representation.

This distinction is fundamental: statistical tokenization is \emph{semantically blind} to the domain because the vocabulary emerges from distributional properties; ontological tokenization is \emph{semantically committed} to a conceptualization of the domain, explicitly inherited from the ontology $\mathcal{O}$.

\subsection{Related Work}

\subsubsection{Subword Tokenization and Its Semantic Impact}

The dominant family of statistical tokenizers --- BPE \citep{Sennrich2016}, WordPiece \citep{Devlin2019}, and SentencePiece \citep{Kudo2018} --- shares a methodological assumption: optimal vocabularies emerge from distributional properties of a corpus, under a compression criterion. Recent studies problematize this assumption on three complementary fronts. \citet{Bostrom2020} show that unigram LM segmentation produces units more aligned with morphology than BPE, indicating that the greedy compression criterion fragments legitimate morphemes. \citet{Rust2021} demonstrate, across nine typologically diverse languages, that a dedicated monolingual tokenizer contributes to monolingual performance as much as the volume of pre-training data, isolating the effect of tokenization from scale. \citet{Schmidt2024} introduce the PathPiece tokenizer and empirically establish that fewer tokens does not imply better downstream performance, dissolving the informal equation between compression and quality. The question of cross-language equity is addressed by \citet{Petrov2023}, who document differences of up to an order of magnitude in tokenization length between languages for informationally equivalent content, with consequences for cost, latency, and effective context window. \citet{Wegmann2025} extend the argument to intralinguistic variation, showing that pre-tokenization decisions interact with orthographic and dialectal variants; in morphologically rich languages \citet{Toraman2023} demonstrate that tokenizer choice affects downstream performance comparably to scale increases, with direct implications for the tokenization of technical PT-BR. \citet{Land2024} further catalogue \emph{glitch tokens} --- tokens present in the vocabulary but virtually absent from training --- as a class of systematic failure induced by the disconnect between tokenizer construction and model training. For \textsc{TOTEN}, this body of evidence is convergent: statistical vocabulary derivation introduces biases, fragmentations, and artifacts that are not corrected a posteriori by the consumer model.

\subsubsection{Numerical Representation and Reasoning in Language Models}

The specific literature on numbers in NLP, synthesized by \citet{Thawani2021} across seven subtasks, identifies numerical representation as a weak and unstable emergent capability in generic models. \citet{Spithourakis2018} show that hierarchical architectures treating numerals as a distinct class reduce perplexity by two to four orders of magnitude on numerical subsets. \citet{Wallace2019} establish, via probing, that standard embeddings capture magnitude only for integers up to three digits, collapsing for larger scales. \citet{Geva2020} propose injecting numerical ability via synthetic arithmetic data generation during pre-training, an approach complementary to the input reformulation explored by \citet{Singh2024}. The limitations documented by \citet{Yang2025} reinforce that the problem persists in frontier models. \textsc{TOTEN} contributes to this discussion from a distinct angle: rather than injecting numeracy via training or redesigning architectures, it operates on the input representation, preserving numerical structure (sign, mantissa, exponent, locale, right-to-left digit grouping per \citealp{Singh2024}) as ontologically typed information before any consumer model.

Structured quantity extraction has a parallel trajectory. \citet{Roy2015} formalize the problem of \emph{Quantity Entailment} and reasoning with quantities in natural language; \citet{Saha2017} present BONIE, the first numerical extractor in Open Information Extraction, inferring implicit relations from contextual cues (e.g., the unit km² suggesting area). \citet{Almasian2023} consolidate this line with CQE, a hybrid system with symbolic and statistical components evaluated on scientific corpora in English. \citet{Zaratiana2024} generalize entity recognition to open-set regime with GLiNER. These systems partially cover the technical space but treat the normative PT-BR vocabulary (NBR, ABNT, hierarchical legal identifiers) as noise or generic naming, without explicit ontological modeling.

\subsubsection{Ontological Engineering and Knowledge-Based Extraction}

Research on \emph{Ontology-Based Information Extraction} (OBIE), systematized by \citet{Wimalasuriya2010}, established architectures in which declared ontologies guide the identification and classification of textual entities. \citet{Maedche2001} propose the ontology learning cycle (import, extract, prune, refine, evaluate), and \citet{Cimiano2006} consolidates methods and metrics in a reference treatise. Upper ontologies such as SUMO \citep{Niles2001} provide a foundation for cross-domain integration. \textsc{TOTEN} inherits from this tradition the epistemic commitment to an explicit formal ontology (OEE), but inverts the typical causal direction: rather than learning an ontology from a corpus, it declares it a priori and classifies textual regions according to primary types whose invariants are verifiable by construction. This inversion is deliberate and compatible with domains in which the ontology already exists institutionally --- engineering has centuries of dimensional, normative, and symbolic codification that pre-exist any particular corpus.

\subsubsection{Knowledge-Based Systems and the Neurosymbolic Debate}

The epistemological defense of systems committed to structure received an influential argument from \citet{Bender2020}, according to which models trained solely on form have no mechanism for learning meaning. \citet{Lake2017}, in a peer-reviewed \textit{target article} of \textit{Behavioral and Brain Sciences}, argue that robust systems require explicit causal and compositional models articulated with statistical learning. The current neurosymbolic synthesis, surveyed by \citet{Hitzler2022} and \citet{Sarker2021}, provides the contemporary framework for hybrid approaches; the modern program of \emph{Inductive Logic Programming} \citep{Cropper2022} illustrates that declarative paradigms continue to evolve methodologically. \textsc{TOTEN} does not compete with neural models: it positions itself as a symbolic pre-processing layer whose output --- a language $\mathcal{M}$ ontologically typed --- can be consumed both by statistical models and by downstream symbolic agents, in an arrangement compatible with the neurosymbolic taxonomy of type \textit{symbolic[neural]} \citep{Sarker2021}.

\subsubsection{Brazilian Portuguese Processing and Benchmarks}

The PT-BR ecosystem has progressively matured: \citet{Souza2020} consolidate the statistical foundation with BERTimbau; the shared tasks ASSIN \citep{Fonseca2016} and ASSIN~2 \citep{Real2020} provide semantic similarity and textual inference datasets; LeNER-Br \citep{LuzDeAraujo2018} illustrates named entity recognition in the Brazilian legal domain with specific classes (legislation, case law). Recent academic benchmarks \citep{Almeida2023bluex,Nunes2023,Almeida2023} enable evaluation in the national domain. None of these resources explicitly models ontological entities in the OEE sense; LeNER-Br approaches this by introducing normative classes, but maintains statistical treatment of mentions. \textsc{TOTEN} is designed to accommodate, in the OEE, the normative-technical vocabulary (NBR, ABNT, compositional unit identifiers, hierarchical references) that remains absent from consolidated PT-BR benchmarks; its validation on an annotated open corpus for that vocabulary is left as an extension (Section~\ref{sec:conclusao}).

The positioning of \textsc{TOTEN} remains, therefore, orthogonal: it does not compete with BPE in compression, with Pint \citep{Grecco2022} or udunits-2 \citep{Hankin2020} in dimensional conversion, with quantitative extractors \citep{Almasian2023,Saha2017} in generic recall, nor with BERTimbau in distributed representation. It acts as an ontological classification layer that consumes external oracles and produces a domain-semantically committed representation, recovering the intrinsic properties of technical entities that purely statistical pipelines systematically lose.

\section{Ontology of Engineering Entities}
\label{sec:oee}

\subsection{Primary Types}

The OEE declares a finite set $\mathcal{T}$ of primary types, each characterized by a signature $\langle \pi_\tau,\ \iota_\tau\rangle$ where $\pi_\tau$ is the set of intrinsic properties and $\iota_\tau$ is the set of invariants that any instance of $\tau$ must preserve. The primary types comprise physical quantities, technical prose, technical identifiers, formal operators, universal constants, structural relations, symbolic expressions, pure numbers, and hierarchical references. This enumeration is closed to the ontology's principles but open to extension as demanded empirically by the domain.

For example, the type \emph{Physical Quantity} has the signature
\begin{equation}
\pi = \langle \mathrm{value},\ \mathrm{unit},\ \mathrm{dim}\rangle,
\end{equation}
where $\mathrm{value} \in \mathbb{R} \cup \{\bot\}$, $\mathrm{unit}$ is a string in a compositional unit language, and $\mathrm{dim} \in \mathbb{Z}^7$ is the dimensional vector in the canonical order of the International System \citep{BIPM2019}. The essential invariant is \emph{dimensional homogeneity}: two instances may be combined by addition only when their dimensional vectors coincide.

\subsection{Structural Principles}

The ontology is governed by eight structural principles that govern recognition, instantiation, and composition. We state here the four central structural axioms ($A_1$, $A_3$, $A_4$, and $A_5$) in their normative form; the complete set $A_1$--$A_8$ is reproduced in Appendix~\ref{app:axiomas}:

\begin{axiomstar}[$A_1$ --- Intrinsicity]
A type $\tau \in \mathcal{T}$ is defined by its intrinsic properties $\pi_\tau$, not by pragmatic criteria or empirical frequency in a corpus.
\end{axiomstar}

\begin{axiomstar}[$A_3$ --- Mediated composition]
For $\tau_1, \tau_2 \in \mathcal{T}$, the composition $\tau_1 \circ \tau_2$ is defined if and only if $(\tau_1, \tau_2) \in \mathcal{R}$. Free concatenation is prohibited.
\end{axiomstar}

\begin{axiomstar}[$A_4$ --- Categorical error]
Applying instantiation $\mathrm{inst}_{\tau'}$ to a region classified as $\tau \neq \tau'$ constitutes a categorical error, not a gradual loss of quality.
\end{axiomstar}

\begin{axiomstar}[$A_5$ --- Closed-for-modification extensibility]
$\mathcal{O}_n$ admits extension to $\mathcal{O}_{n+1}$ provided $\mathcal{T}_n \subseteq \mathcal{T}_{n+1}$, $\mathcal{P}_n \subseteq \mathcal{P}_{n+1}$, and no invariant in $\mathcal{I}_n$ is violated by $\mathcal{O}_{n+1}$.
\end{axiomstar}

The four remaining principles --- $A_2$ (invariant preservation by valid representation), $A_6$ (typographic convention as intrinsic property of notation), $A_7$ (structural anchoring of symbolic expressions in adjacent formal operators), and $A_8$ (distinctive mathematical mark in every compound symbol) --- are stated in complete normative form in Appendix~\ref{app:axiomas}.

\subsection{Composition Relations}

The relation $\mathcal{R}$ is defined explicitly by enumeration. Notably, $(\mathrm{UniversalConstant},\ \mathrm{PhysicalQuantity}) \in \mathcal{R}$: every named universal constant composes a physical quantity with an SI unit (e.g., $k_B = 1.38 \times 10^{-23}$~J/K). Analogously, $(\mathrm{SymbolicExpression},\ \mathrm{FormalOperator}) \in \mathcal{R}$: every symbolic expression in prose requires anchoring by an adjacent relational or calculus operator.

\section{TOTEN Architecture}
\label{sec:arquitetura}

\begin{figure*}[t]
\centering
\includegraphics[width=.9\textwidth]{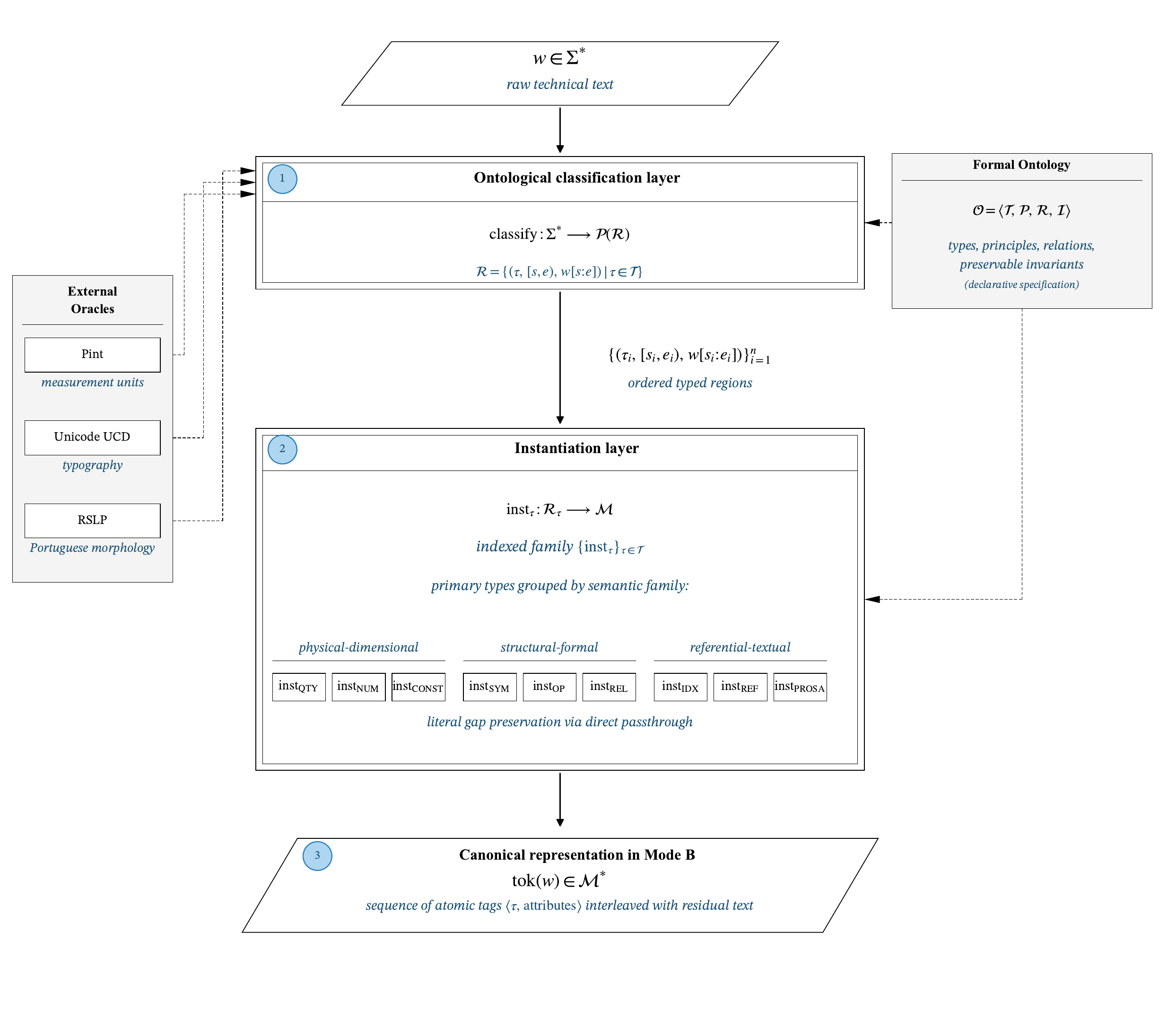}
\caption{Architecture of \textsc{TOTEN.} The ontological classification layer maps raw text into typed regions by consulting the three consolidated external oracles (Pint, Unicode Character Database, RSLP) and the declarative specification of the OEE ontology. The instantiation layer comprises an indexed family of functions, one per type, producing the structured representation in Mode B.}
\label{fig:arquitetura}
\end{figure*}

\textsc{TOTEN} is an operational instantiation of the OEE ontology in three functional layers, coupled to three consolidated external oracles. Figure~\ref{fig:arquitetura} summarizes the architecture.

\subsection{Ontological Classification Layer}

The classification layer is a total function
\begin{equation}
\mathrm{classify}: \Sigma^* \longrightarrow \mathcal{P}(\mathcal{R}_{\Sigma}),
\label{eq:classify}
\end{equation}
where
\begin{equation}
\mathcal{R}_{\Sigma} = \{(\tau,\ [s, e),\ w[s\!:\!e])\ \mid\ \tau \in \mathcal{T},\ 0 \leq s < e \leq |w|\}
\end{equation}
is the set of typed regions in $w \in \Sigma^*$. A region $(\tau, [s, e), c)$ associates a type $\tau$, a position interval $[s, e) \subset [0, |w|)$, and the literal content $c = w[s\!:\!e]$. The image of $\mathrm{classify}(w)$ is a set linearly ordered by starting position, with overlap resolution determined by a precedence relation $\succ_{\mathcal{T}}$ declared in $\mathcal{O}$. Monotonicity of the function with respect to substring inclusion is preserved: for all $w' \sqsubseteq w$, $\mathrm{classify}(w') \subseteq \mathrm{classify}(w) \cap \mathcal{R}_{w'}$.

\subsection{Instantiator Family}

The instantiation layer is the indexed family $\{\mathrm{inst}_\tau\}_{\tau \in \mathcal{T}}$ where each component
\begin{equation}
\mathrm{inst}_\tau:\ \mathcal{R}_\tau \longrightarrow \mathcal{M}
\label{eq:inst}
\end{equation}
maps regions of type $\tau$ into strings of the output language $\mathcal{M}$. The composition $\mathrm{inst}_\tau \circ \mathrm{classify}|_{\mathcal{R}_\tau}$ produces the ordered sequence of type-$\tau$ tags corresponding to a text $w$. The final concatenated result, interleaved with unclassified residual text, constitutes the representation $\mathrm{tok}(w) \in \mathcal{M}^*$.

The categorical separation between classification and instantiation implements the \emph{single authority} principle: the classification layer decides the type of each region; the instantiation layer does not re-decide, it only formats. Type errors, per Axiom~$A_4$ (Categorical error), propagate as categorical exceptions.

\subsection{Coupling with External Oracles}

The ontological robustness of the system derives from coupling with three consolidated external oracles, replacing manual enumeration of cases by delegation to established authorities.

\emph{Dimensional domain.} The \textit{Pint} library \citep{Grecco2022} is the external authority on units of measure. Dimensional atoms are materialized deterministically from Pint's unit registry, with expansion by International System prefixes, validation of positive conversion factors, and exclusion of physical constants that belong to another ontological type. Dimensional composition over the $\mathbb{Z}^7$ vector is delegated to Pint.

\emph{Typographic domain.} The \textit{Unicode Character Database} \citep{Unicode2024} is queried to identify typographic markers without character-by-character enumeration. Portuguese ordinals are identified by the decomposition property of type \textit{super} combined with a specific Latin letter; numeric superscripts by the same property applied to digits; mathematical operators by the general category \emph{Sm}.

\emph{Morphological domain.} The \emph{RSLP} algorithm \citep{Orengo2001}, the established standard for Portuguese morphology, reduces any gender, number, or derivational inflection to the lemmatic root. \textsc{TOTEN} employs it to detect contextual technical anchors associated with single-ASCII-letter units, allowing occurrences such as \textit{temperatura} (temperature), \textit{tens\~{a}o} (tension), or \textit{pot\^{e}ncias} (powers) to confirm the technical use of an ambiguous unit without the need to manually enumerate all Portuguese inflections.

\section{Output Language}
\label{sec:representacao}

The output language $\mathcal{M}$ is defined over an extended alphabet $\Sigma \cup \mathcal{D}$, where $\mathcal{D}$ is a set of structural delimiters. The production in Backus-Naur Form (BNF) is
\begin{align}
\mathrm{tag}       &\to \texttt{[} \tau\ \mathrm{attributes} \texttt{]} \\
\mathrm{attributes}&\to \mathrm{attribute} \mid \mathrm{attribute}\ \mathrm{attributes} \\
\mathrm{attribute} &\to \mathrm{key} \texttt{=} \mathrm{value}
\end{align}
where $\tau \in \mathcal{T}$ identifies the type, $\mathrm{key}$ is an alphanumeric identifier, and $\mathrm{value}$ is a quoted string, a normalized number, or an integer vector. The complete representation $\mathrm{tok}(w) \in \mathcal{M}^*$ is the ordered concatenation of tags with the unclassified residual text between them, preserved verbatim. This property --- \emph{literal preservation} of the source text outside typed regions --- distinguishes $\mathcal{M}$ from annotation markup languages typically used in linguistic corpora, in which the source text is replaced or rewritten.

Each type $\tau$ defines a signature $\Pi_\tau$ of mandatory and optional attributes. For physical quantity, the mandatory attributes are $\{\mathrm{value},\allowbreak \mathrm{unit},\allowbreak \mathrm{dim}\}$ and the optional ones are $\{\mathrm{r2l},\allowbreak \mathrm{ambig},\allowbreak \mathrm{alternatives}\}$. For pure number, the mandatory attributes are $\{\mathrm{value},\allowbreak \mathrm{locale},\allowbreak \mathrm{repr},\allowbreak \mathrm{original}\}$. For hierarchical reference, $\mathrm{hierarchy}$ is mandatory. For technical identifier, $\mathrm{slug}$ is mandatory. The optional attributes $\mathrm{ambig}$ and $\mathrm{alternatives}$ are reserved for a future contextual ambiguity resolution layer (see Conclusion); they are not exercised in this study.

The invariance of \emph{value} under alternative IEEE 754 representations is guaranteed by deterministic canonicalization that maps the original number string to a unique \textit{float}, modulo machine precision. The \texttt{original} attribute in number tags preserves the exact form written by the author, maintaining locale (Brazilian or English), thousands separator, and representation (decimal, scientific, fractional, percentage, ordinal).

\section{Intrinsic Evaluation}
\label{sec:avaliacao}

The evaluation adopts an intrinsic protocol based on four properties verifiable by construction, formally defined below.

\subsection{Verifiable Properties}

Let $S$ be an evaluated tokenization system, $G$ the set of entities annotated in the \textit{ground truth} of a corpus $\mathcal{C}$, and $S(g)$ the representation produced by $S$ for entity $g \in G$.

\begin{definition}[Ontological atomicity]
\label{def:h1}
Property $H_1(S, g)$ is true if and only if $S(g)$ is a single indivisible tag corresponding to the correct type of $g$. Formally,
\begin{equation}
H_1(S, g) = 1 \;\Leftrightarrow\; |S(g)| = 1 \;\wedge\; \tau(S(g)) = \tau(g).
\end{equation}
\end{definition}

\begin{definition}[Dimensional equivalence]
For a pair $(g_1, g_2)$ of dimensionally equivalent physical quantities,
\begin{equation}
H_2(S, g_1, g_2) = 1 \;\Leftrightarrow\; \mathrm{dim}(S(g_1)) = \mathrm{dim}(S(g_2)),
\end{equation}
with equality in $\mathbb{Z}^7$.
\end{definition}

\begin{definition}[Typographic robustness]
For a group $\mathcal{V}$ of semantically equivalent notational variants,
\begin{equation}
H_3(S, \mathcal{V}) = 1 \;\Leftrightarrow\; \forall v, v' \in \mathcal{V}:\ \tau(S(v)) = \tau(S(v')).
\end{equation}
\end{definition}

\begin{definition}[Numerical reconstruction]
$H_4(S, g) = 1$ if and only if the pair $(\mathrm{value}(S(g)),\allowbreak \mathrm{unit}(S(g)))$ is programmatically extractable and satisfies
\begin{align}
&|\mathrm{value}(S(g)) - \mathrm{value}(g)| < \varepsilon, \\
&\mathrm{dim}(\mathrm{unit}(S(g))) = \mathrm{dim}(\mathrm{unit}(g)),
\end{align}
with fixed tolerance $\varepsilon = 10^{-6}$ (absolute error over the IEEE~754 canonicalized value).
\end{definition}

\subsection{Statistical Metrics}

For binary per-instance hypotheses, paired contrasts between systems are evaluated by the McNemar test with exact computation \citep{Dietterich1998}. Confidence intervals for proportions use the \citet{Wilson1927} formula. Effect size between paired proportions is quantified by Cohen's $h$ coefficient \citep{Cohen1988}. Correction for multiple comparisons follows the \citet{Holm1979} procedure.

\subsection{Corpora}

The internal corpus is the \textit{EngQuant} benchmark, with $N = 800$ cases generated procedurally over five structural typologies (cantilever beam, simply supported beam, simple plane frame, truss, and element under combined load), with physical validation via the OpenSeesPy simulator \citep{Zhu2018}. Compositional diversity covers seven independent generation dimensions.

Cross-corpus validation employs four external corpora in Brazilian Portuguese: \textit{MMMLU PT\_BR}, a professional translation of the MMLU benchmark \citep{Hendrycks2021} by OpenAI, with 595 cases eligible for numerical reconstruction; \textit{BLUEX} \citep{Almeida2023bluex}, aggregating USP and UNICAMP entrance examinations from 2018 to 2023, with 151 eligible cases; \textit{ENEM Maritaca}, with 83 eligible cases from the National High School Exam (ENEM) from 2022 to 2024; and \textit{Alvorada-Bench} \citep{Nunes2023}, aggregating FUVEST, IME, and ITA, with 942 eligible cases.

\subsection{Comparative Systems}

We compare \textsc{TOTEN} against eight representative systems in three families. Statistical tokenizers: \textit{cl100k} and \textit{o200k} \citep{Brown2020}. Specialized quantitative extractors: Quantulum3, CQE \citep{Almasian2023}, and GLiNER \citep{Zaratiana2024}. Dimensional libraries: Pint \citep{Grecco2022} and udunits-2 \citep{Hankin2020}. Generic entity recognition in Portuguese: spaCy, model \textit{pt-core-news-md} \citep{Honnibal2020}.

\section{Results}
\label{sec:resultados}

\begin{figure*}[t]
\centering
\includegraphics[width=.9\textwidth]{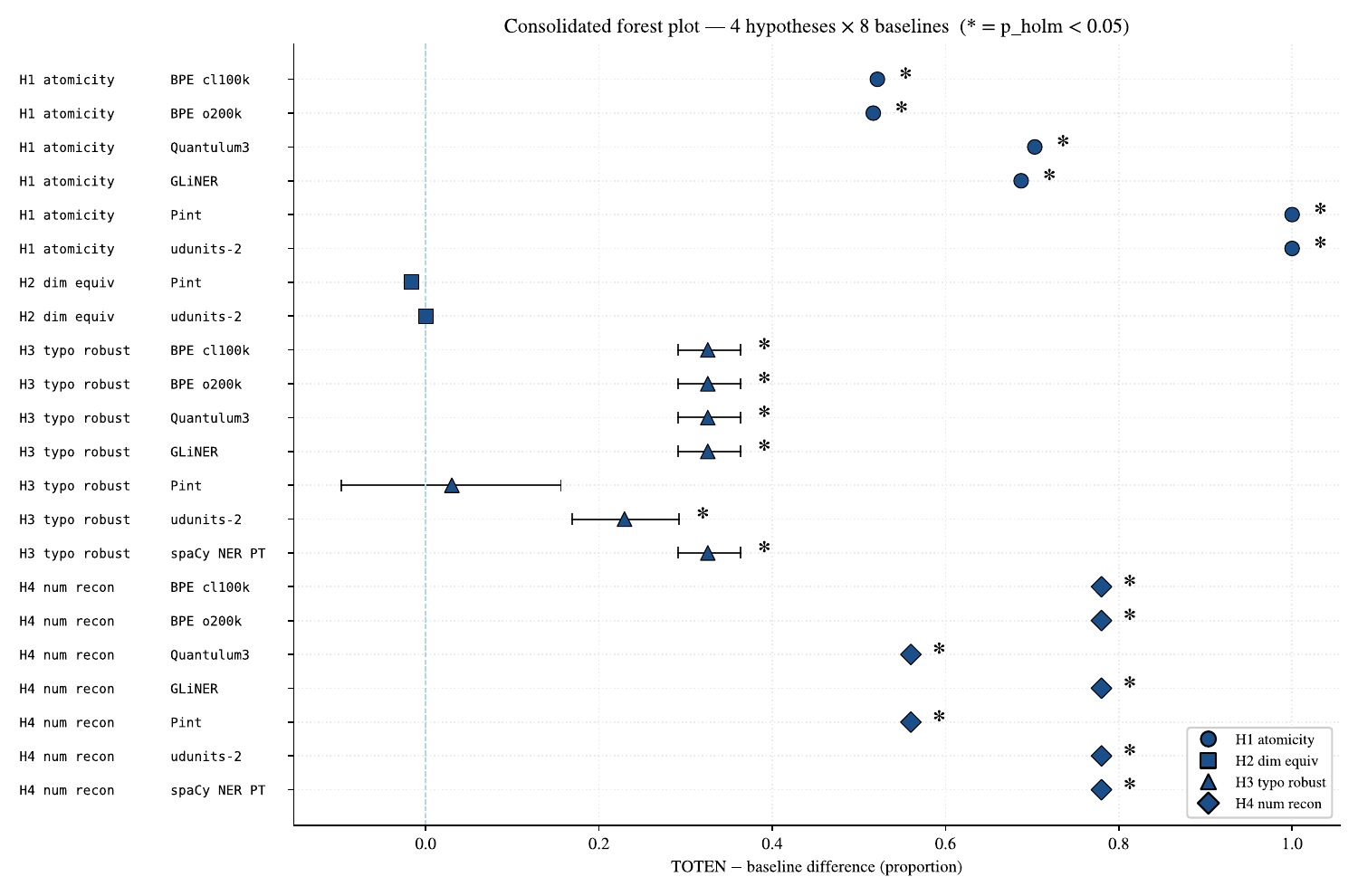}
\caption{Consolidated summary of paired contrasts on the internal benchmark. Points represent the proportion difference $S_{\text{TOTEN}} - S_{\text{baseline}}$; bars represent 95\% Wilson confidence intervals. Differences in $H_1$ and $H_4$ are statistically significant by McNemar with Holm correction in all contrasts ($p < 0.001$).}
\label{fig:summary-forest}
\end{figure*}

Figure~\ref{fig:summary-forest} summarizes the consolidated contrasts between \textsc{TOTEN} and each comparative system on the internal benchmark. Table~\ref{tab:resultados-principais} presents absolute values per property and system.

\begin{table*}[ht]
\caption{Consolidated results on the internal EngQuant benchmark.\label{tab:resultados-principais}}
\begin{tabular*}{\tblwidth}{@{\extracolsep{\fill}} l c c c c @{}}
\toprule
\textbf{System} & \textbf{$H_1$ atomicity} & \textbf{$H_2$ equivalence} & \textbf{$H_3$ typography} & \textbf{$H_4$ reconstruction} \\
\midrule
\textbf{TOTEN}                & \textbf{$1.000$} & \textbf{$0.968$} & \textbf{$0.326$} & \textbf{$0.780$} \\
BPE \texttt{cl100k\_base}     & $0.479$          & ---              & $0.000$          & $0.000$          \\
BPE \texttt{o200k\_base}      & $0.483$          & ---              & $0.000$          & $0.000$          \\
Quantulum3                    & $0.297$          & ---              & $0.000$          & $0.220$          \\
CQE                           & $0.225$          & ---              & $0.000$          & $0.340$          \\
GLiNER                        & $0.313$          & ---              & $0.000$          & $0.000$          \\
Pint \textit{standalone}      & $0.000$          & \textbf{$0.985$} & $0.295$          & $0.220$          \\
\textit{udunits-2}            & $0.000$          & $1.000$          & $0.096$          & $0.000$          \\
spaCy NER pt                  & $0.257$          & ---              & $0.000$          & $0.000$          \\
\bottomrule
\end{tabular*}
\end{table*}

\subsection{Detection and Ontological Atomicity}

\begin{figure*}[t]
\centering
\includegraphics[width=.9\textwidth]{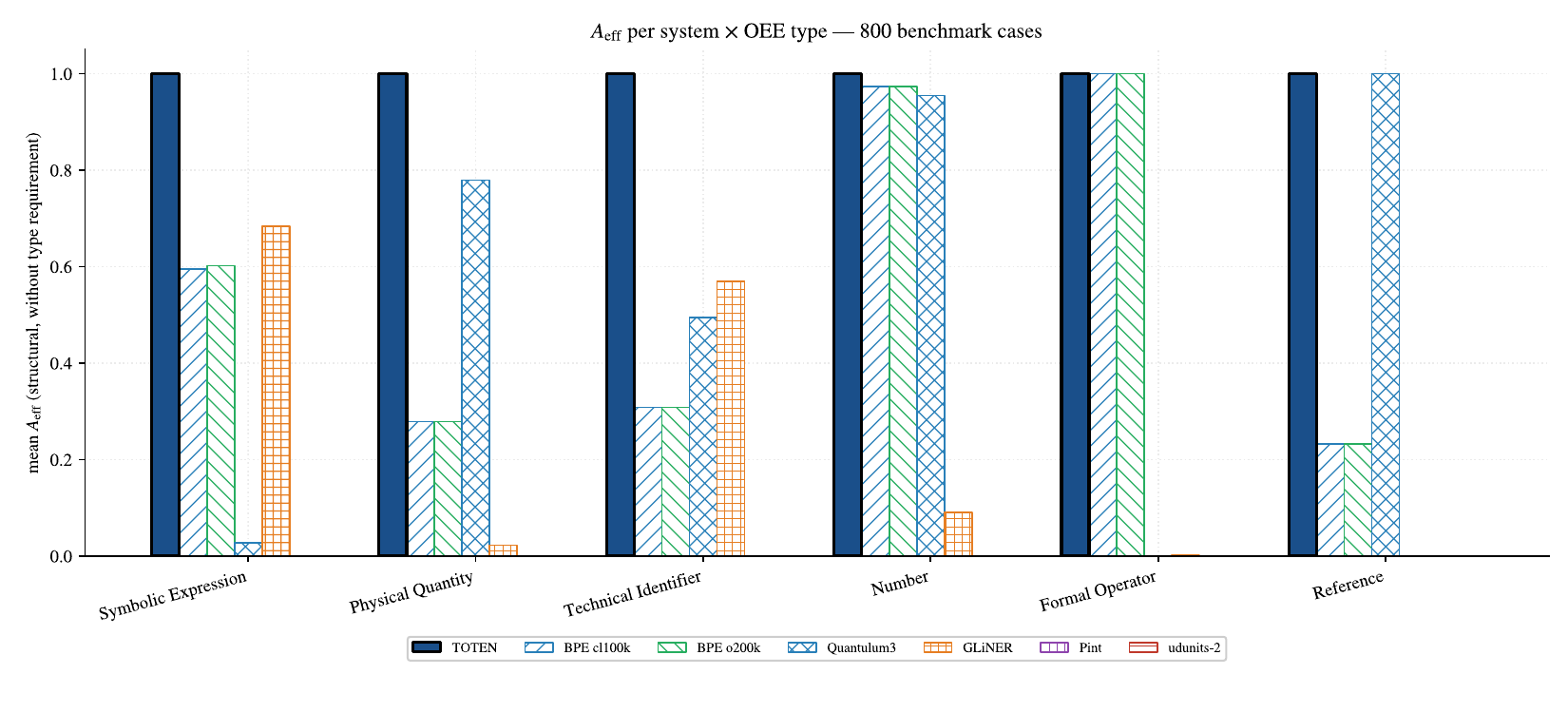}
\caption{Atomicity by system and ontological type. \textsc{TOTEN} is the only system that combines unit \emph{recall} and atomicity across all evaluated types (physical quantity, technical identifier, formal operator, symbolic expression, and number).}
\label{fig:h1}
\end{figure*}

\textsc{TOTEN} achieves unit atomicity over the $|G| = 31\,674$ annotated ground-truth entities evaluated. To avoid a tautological reading of $H_1 = 1.000$, we decompose recognition into three metrics (Table~\ref{tab:recall_h1}): detection \emph{recall} ($|R_S|/|G|$, where $R_S$ is the set recognized by system $S$ and $G$ is the ground truth), conditional structural atomicity ($A_{\mathrm{cond}}$, the fraction of recognized regions emitted as a single indivisible tag, \emph{without} type requirement), and effective structural atomicity ($A_{\mathrm{eff}} = \mathrm{Recall} \cdot A_{\mathrm{cond}}$, which penalizes non-detection). Ontological atomicity $H_1$ (Definition~\ref{def:h1}, Table~\ref{tab:resultados-principais}) is stricter: it additionally requires that the single tag receive the correct type, so $H_1 \leq A_{\mathrm{eff}}$. The separation between detection and correct classification is the standard evaluative convention in ontology-based extraction \citep{Wimalasuriya2010}. BPE tokenizers have unit recall but emit a single tag for about two-thirds of entities ($A_{\mathrm{eff}} \approx 0.66$), satisfying $H_1$ --- with correct type --- for roughly half (Table~\ref{tab:resultados-principais}); dimensional libraries have zero textual recall (they operate on isolated unit strings); specialized extractors show partial recall. \textsc{TOTEN} combines unit recall over the declared OEE closure with unit structural and ontological atomicity. The advantage is categorical: recall \emph{and} atomicity simultaneously. Entities outside the OEE closure are not recognized by construction, in conformance with Axiom~$A_1$ (Intrinsicity). Figure~\ref{fig:h1} stratifies the result by ontological type.

\begin{table*}[ht]
\caption{Detection recall and structural atomicity on the internal EngQuant benchmark.\label{tab:recall_h1}}
\begin{tabular*}{\tblwidth}{@{\extracolsep{\fill}} l c c c @{}}
\toprule
\textbf{System} & \textbf{Recall} & \textbf{$A_{\mathrm{cond}}$} & \textbf{$A_{\mathrm{eff}}$} \\
\midrule
\textbf{TOTEN}           & \textbf{$1.000$ $[1.000,\ 1.000]$} & \textbf{$1.000$ $[1.000,\ 1.000]$} & \textbf{$1.000$ $[1.000,\ 1.000]$} \\
BPE \texttt{cl100k}      & $1.000$ $[1.000,\ 1.000]$ & $0.655$ $[0.650,\ 0.661]$ & $0.655$ $[0.650,\ 0.661]$ \\
BPE \texttt{o200k}       & $1.000$ $[1.000,\ 1.000]$ & $0.658$ $[0.653,\ 0.663]$ & $0.658$ $[0.653,\ 0.663]$ \\
Quantulum3               & $0.390$ $[0.384,\ 0.395]$ & $0.881$ $[0.875,\ 0.887]$ & $0.343$ $[0.338,\ 0.349]$ \\
CQE                      & $0.230$ $[0.225,\ 0.235]$ & $0.988$ $[0.985,\ 0.990]$ & $0.227$ $[0.223,\ 0.232]$ \\
GLiNER                   & $0.313$ $[0.308,\ 0.318]$ & $1.000$ $[0.999,\ 1.000]$ & $0.313$ $[0.308,\ 0.318]$ \\
Pint \textit{standalone} & $0.000$ $[0.000,\ 0.000]$ & ---                        & $0.000$ $[0.000,\ 0.000]$ \\
\textit{udunits-2}       & $0.000$ $[0.000,\ 0.000]$ & ---                        & $0.000$ $[0.000,\ 0.000]$ \\
spaCy NER pt             & $0.259$ $[0.254,\ 0.264]$ & $0.997$ $[0.996,\ 0.998]$ & $0.258$ $[0.253,\ 0.263]$ \\
\bottomrule
\end{tabular*}
\end{table*}

\subsection{Dimensional Equivalence}

\begin{figure*}[t]
\centering
\includegraphics[width=.9\textwidth]{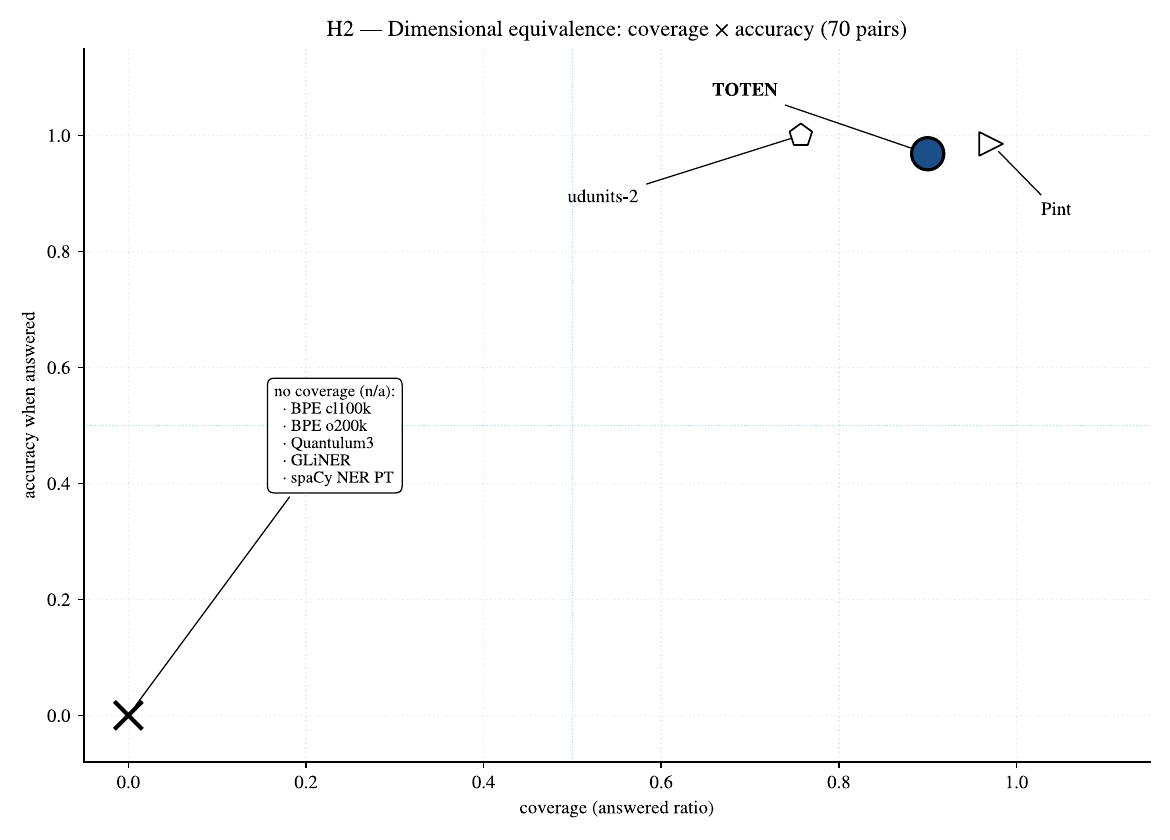}
\caption{Coverage and accuracy in dimensional equivalence. \textsc{TOTEN} balances coverage and accuracy, with a non-significant difference relative to Pint.}
\label{fig:h2}
\end{figure*}

\textsc{TOTEN} achieves conditional dimensional accuracy of $0.968$ ($61$ correct answers in $63$ answered pairs, among the $70$ dimensional pairs evaluated), against $0.985$ for Pint over the same sample. The difference $\Delta = -0.017$ is not statistically significant by the McNemar test ($p = 1.0$). The interpretation is direct: \textsc{TOTEN} consumes Pint as a dimensional oracle and therefore inherits its authority. The residual difference reflects operation over continuous text (TOTEN) versus isolated unit strings (Pint). We report $H_2$ in two conventions: $H_2^{\mathrm{cond}}$, conditional on pairs with a valid response (the value $0.968$ in Table~\ref{tab:resultados-principais}), and $H_2^{\mathrm{eff}}$, effective over all $70$ dimensional pairs with non-response counted as error ($61/70 = 0.871$ for the full configuration in the ablation of Table~\ref{tab:ablacao_loo}); the difference between the two forms reflects only coverage, not conditional dimensional accuracy. Figure~\ref{fig:h2} shows the relationship between coverage and conditional accuracy for the three systems with an explicit dimensional vector.

\subsection{Typographic Robustness}

\begin{figure*}[t]
\centering
\includegraphics[width=.9\textwidth]{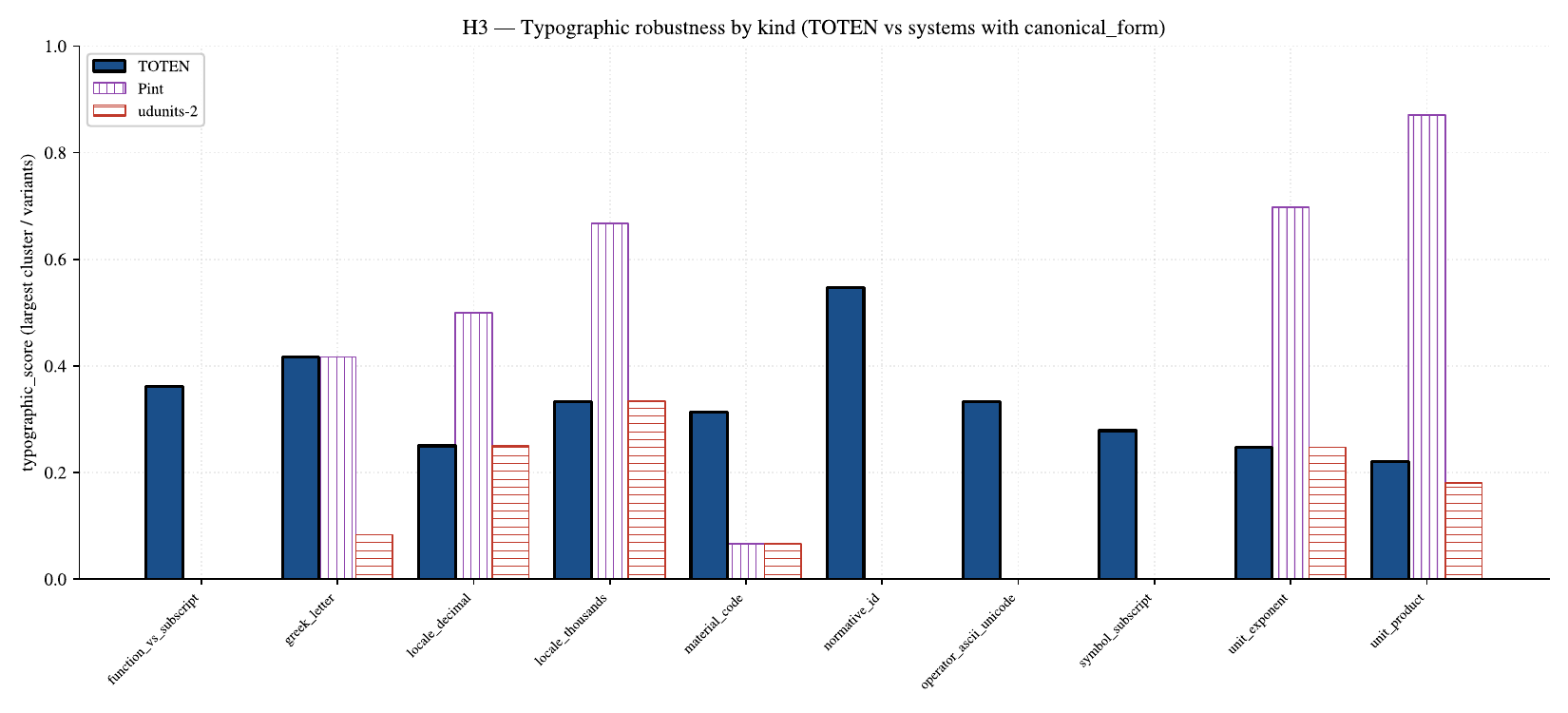}
\caption{Typographic robustness stratified by variant type. For each category of notational variant (e.g., Unicode superscript, PT-BR decimal separator), metric $H_3$ quantifies the fraction of groups whose variants receive an identical type after tokenization.}
\label{fig:h3}
\end{figure*}

Figure~\ref{fig:h3} presents robustness stratified by variant type. We report $H_3$ in two forms (Table~\ref{tab:h3_dual}): $H_3^{\mathrm{global}}$ over all 43 variant groups in the benchmark and $H_3^{\mathrm{scoped}}$ over the 42 groups within the closure declared \emph{a priori} by the OEE coupled to the oracles. \textsc{TOTEN} achieves $H_3^{\mathrm{global}} = 0.326$ (vs.\ $H_3^{\mathrm{scoped}} = 0.325$; removing the single \texttt{locale\_thousands} group deferred to a future phase changes the result by $-0.0002$), particularly on composition with centered dot and compositional space per \citep{BIPM2019}, and on multiple pressure-by-liquid-column variants. Pint achieves $0.295$ ($0.287$ scoped); remaining systems fragment or ignore the entity at first contact with a notational variant not exactly anticipated. The absolute value $0.326$ reflects that typographic coverage is limited to the declared OEE closure coupled to the oracles and Unicode UCD delegation: variants outside this closure are not normalized by construction, and expanding the catalogue is incremental work. The relative advantage over comparatives is preserved under both denominators. The 43 variant groups are distributed across the notational categories of Figure~\ref{fig:h3} (e.g., Unicode superscript, decimal separator, compositional centered dot, pressure variants); each category aggregates multiple semantically equivalent groups.

\begin{table*}[ht]
\caption{Typographic robustness in two evaluation forms.\label{tab:h3_dual}}
\begin{tabular*}{\tblwidth}{@{\extracolsep{\fill}} l c c @{}}
\toprule
\textbf{System} & \textbf{$H_3^{\mathrm{global}}$ ($n=43$)} & \textbf{$H_3^{\mathrm{scoped}}$ ($n=42$)} \\
\midrule
\textbf{TOTEN}           & \textbf{$0.326$ $[0.291,\ 0.364]$} & \textbf{$0.325$ $[0.289,\ 0.365]$} \\
Pint \textit{standalone} & $0.295$ $[0.188,\ 0.408]$ & $0.287$ $[0.181,\ 0.402]$ \\
\textit{udunits-2}       & $0.096$ $[0.058,\ 0.136]$ & $0.090$ $[0.054,\ 0.130]$ \\
\bottomrule
\end{tabular*}
\end{table*}

\subsection{Numerical Reconstruction}

\begin{figure*}[t]
\centering
\includegraphics[width=.9\textwidth]{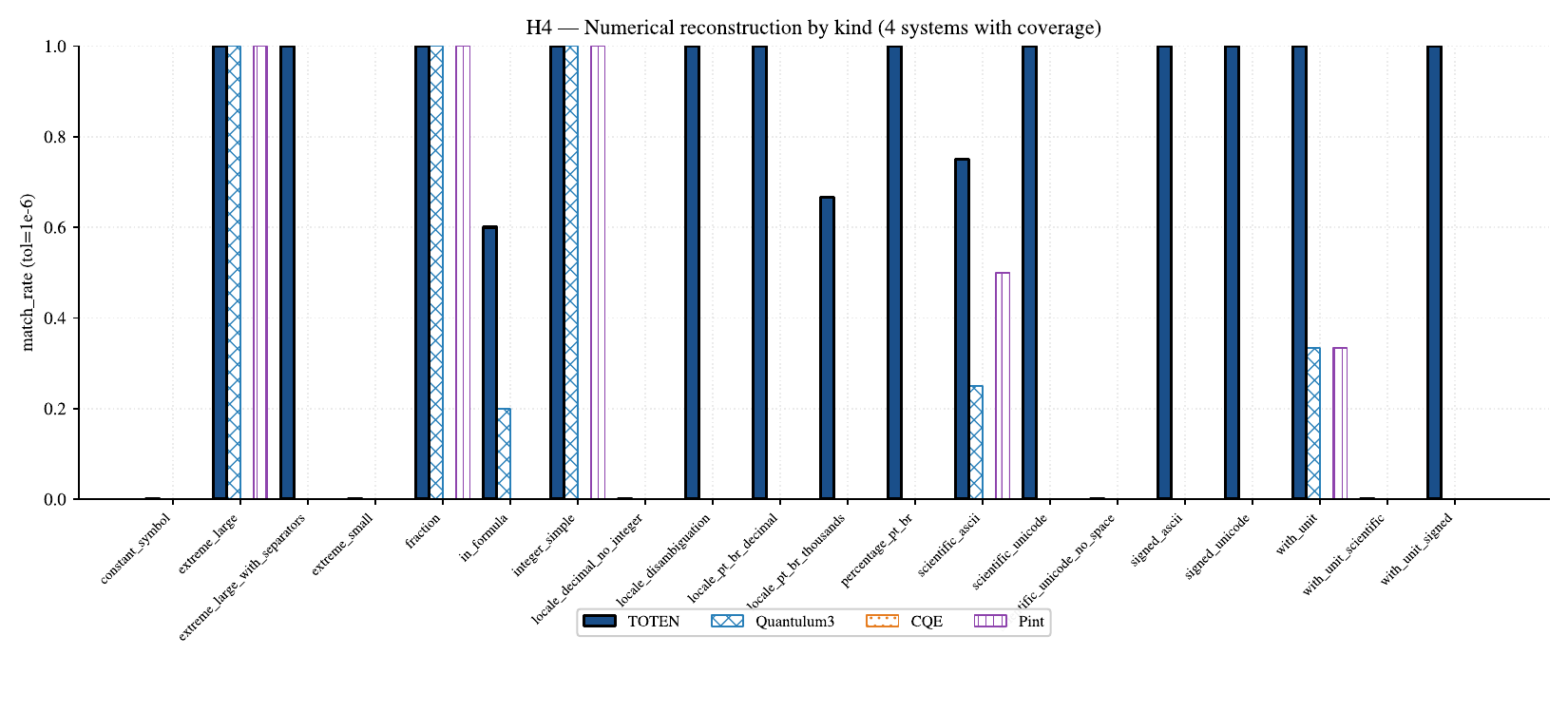}
\caption{Numerical reconstruction stratified by subtype. \textsc{TOTEN} leads or ties in the majority of evaluated numerical subtypes (wins in $11$ and ties in $8$ of the $20$ subtypes), with perfect accuracy on fractions, percentages, PT-BR locale decimals, and Unicode scientific notation; comparative systems cover only partial subsets, with heterogeneous performance across subtypes.}
\label{fig:h4}
\end{figure*}

The most discriminative property of the system. \textsc{TOTEN} achieves $0.780$ on the internal benchmark, against $0.340$ for CQE --- the best internal baseline --- and $0.220$ for Quantulum3 and Pint. Systems without explicit quantitative extraction achieve zero. The difference relative to CQE, $\Delta = +0.440$, is statistically significant by the McNemar test with Holm correction ($p < 10^{-4}$). Figure~\ref{fig:h4} stratifies the result by numerical subtype.

\subsection{Validation on External Corpora}

\begin{figure*}[t]
\centering
\includegraphics[width=.9\textwidth]{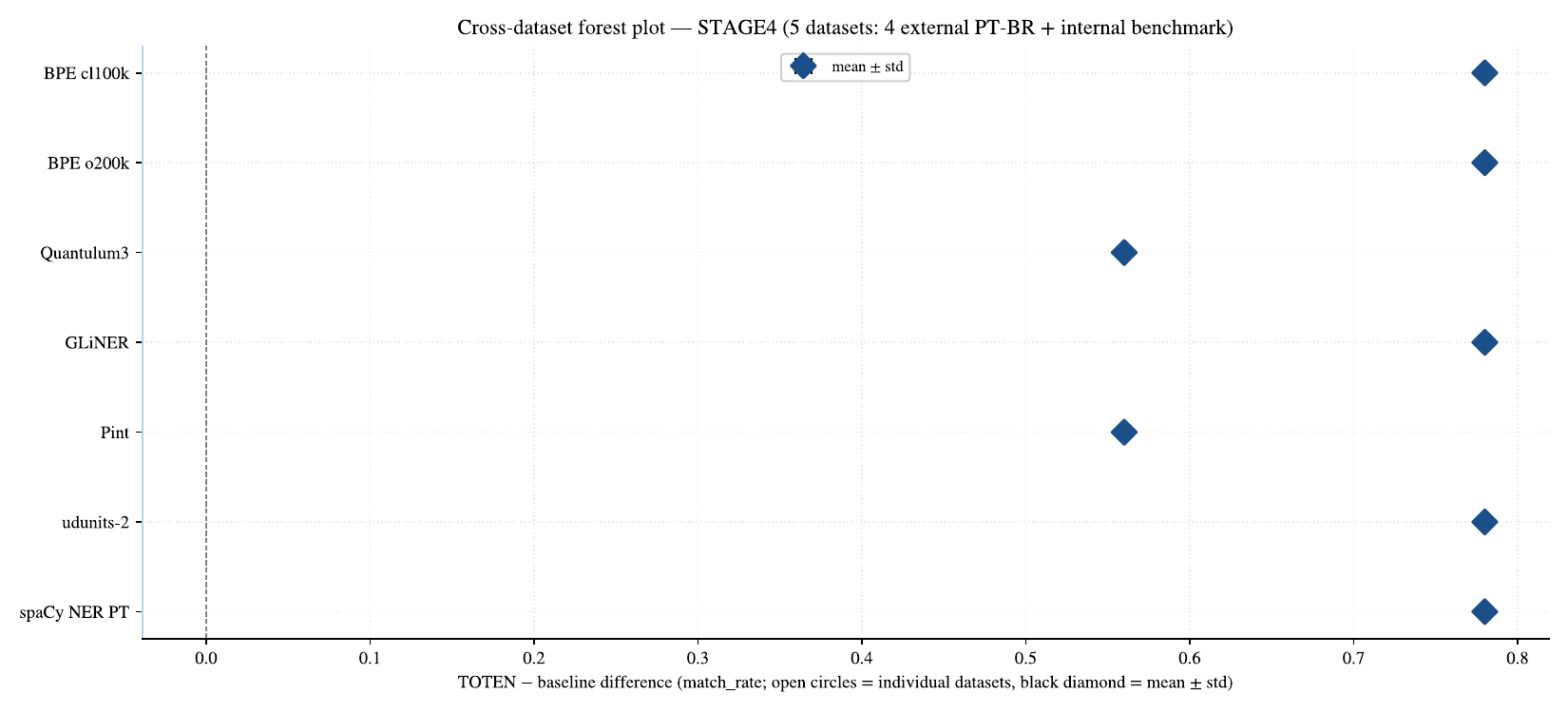}
\caption{Numerical reconstruction on four external Brazilian Portuguese corpora. \textsc{TOTEN} leads in all, with differences significant by McNemar with Holm correction.}
\label{fig:cross-h4}
\end{figure*}

\begin{figure*}[t]
\centering
\includegraphics[width=.9\textwidth]{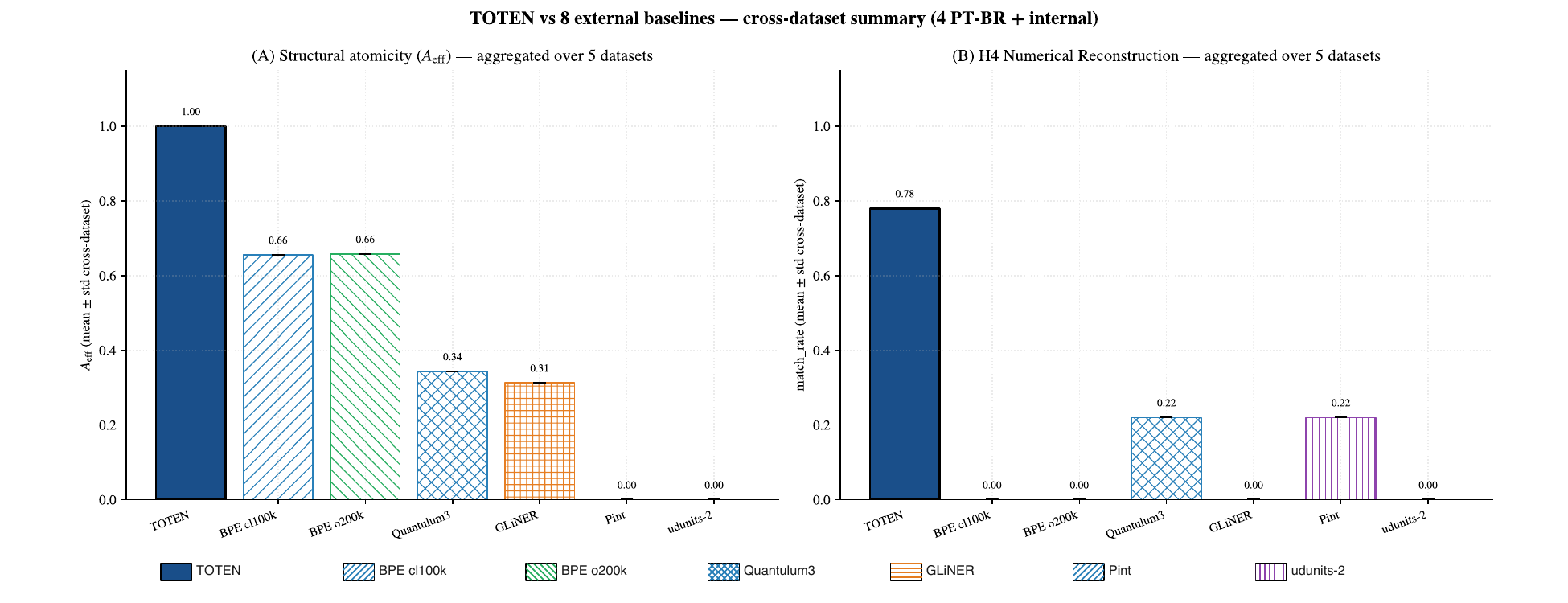}
\caption{Comparative synthesis on the internal EngQuant benchmark. Panel (A): effective structural atomicity ($A_{\mathrm{eff}}$, which does not require correct type; cf.\ Table~\ref{tab:recall_h1}); Panel (B): numerical reconstruction ($H_4$). \textsc{TOTEN} maintains unit or near-unit atomicity and reconstruction against all comparative systems.}
\label{fig:thesis-card}
\end{figure*}

Table~\ref{tab:externos} summarizes numerical reconstruction results on the four external Brazilian Portuguese corpora. \textsc{TOTEN} maintains leadership across all: $0.866$ on MMMLU PT\_BR, $0.775$ on BLUEX, $0.904$ on ENEM Maritaca, and $0.790$ on Alvorada-Bench. The best comparative system, Quantulum3, achieves between $0.627$ and $0.703$. Figure~\ref{fig:cross-h4} presents the consolidated forest plot of contrasts on external corpora. Figure~\ref{fig:thesis-card} consolidates the final cross-corpus synthesis.

\begin{table*}[ht]
\caption{Numerical reconstruction on external Brazilian Portuguese corpora.\label{tab:externos}}
\begin{tabular*}{\tblwidth}{@{\extracolsep{\fill}} l c c c c @{}}
\toprule
\textbf{System} & \textbf{MMMLU PT\_BR} & \textbf{BLUEX} & \textbf{ENEM Maritaca} & \textbf{Alvorada} \\
                & $n=595$ & $n=151$ & $n=83$ & $n=942$ \\
\midrule
\textbf{TOTEN}           & \textbf{$0.866$} & \textbf{$0.775$} & \textbf{$0.904$} & \textbf{$0.790$} \\
Quantulum3               & $0.703$          & $0.662$          & $0.627$          & $0.645$          \\
CQE                      & $0.523$          & $0.106$          & $0.590$          & $0.369$          \\
Pint \textit{standalone} & $0.000$          & $0.000$          & $0.012$          & $0.000$          \\
Remaining systems        & $0.000$          & $0.000$          & $0.000$          & $0.000$          \\
\bottomrule
\end{tabular*}
\end{table*}

\subsection{Oracle Ablation}
\label{sec:ablacao}

Table~\ref{tab:ablacao_loo} reports the \emph{leave-one-oracle-out} ablation on the internal EngQuant: for each of the three external oracles coupled to \textsc{TOTEN} --- Pint, Unicode UCD, and RSLP --- we disable the oracle, keep the rest of the ontological architecture, and re-measure the five metrics. Removing Pint produces the most expressive drop, concentrated in Recall ($\Delta = -0.188$; paired McNemar $b=5\,955$, $c=0$, $p<10^{-3}$) and in $H_2$ ($\Delta = -0.871$; $p<10^{-3}$): without the dimensional oracle no unit atoms exist in the recognizer, so physical quantities cease to be detected as atomic entities. The drop in $H_4$ is smaller ($\Delta = -0.040$; $p=0.5$), because the OEE Number type captures pure numerical values independently of dimensional recognition, preserving most of the reconstruction fidelity. Removing UCD produces a residual effect on the internal benchmark ($\Delta\,\mathrm{Recall} = -0.003$, $p<10^{-3}$; $\Delta H_3 = 0.000$, $p=1.0$): the contribution of UCD is localized to Unicode mathematical operators ($\approx$, $\leq$, $\geq$, $\times$) and typographic exclusion of \emph{super}/\emph{sub} letters, events infrequent in the structural engineering corpus evaluated. Removing RSLP is \emph{invisible} in this benchmark ($\Delta = 0$ in all metrics; $p=1.0$): EngQuant does not contain regions where single-letter units (K, A, V, W, \ldots) appear ambiguously, so the RSLP technical anchor function is never invoked. This nullity is honest and expected: the contribution of RSLP materializes in corpora where such ambiguities exist, whose domain-specific validation is left as future work.

\begin{table*}[ht]
\caption{Leave-one-oracle-out ablation on the internal EngQuant benchmark.\label{tab:ablacao_loo}}
\footnotesize
\begin{tabular*}{\tblwidth}{@{\extracolsep{\fill}} l c c c c c @{}}
\toprule
\textbf{Configuration} & \textbf{Recall} & \textbf{$H_1$} & \textbf{$H_2$} & \textbf{$H_3$} & \textbf{$H_4$} \\
\midrule
\textbf{TOTEN (full)} & \textbf{$1.000$ $[1.000,\ 1.000]$} & \textbf{$1.000$ $[1.000,\ 1.000]$} & \textbf{$0.871$ $[0.773,\ 0.931]$} & \textbf{$0.326$ $[0.291,\ 0.363]$} & \textbf{$0.780$ $[0.648,\ 0.872]$} \\
TOTEN $-$ Pint  & \textbf{$0.812$ $[0.808,\ 0.816]$} & \textbf{$0.962$ $[0.960,\ 0.964]$} & \textbf{$0.000$ $[0.000,\ 0.052]$} & \textbf{$0.326$ $[0.291,\ 0.363]$} & \textbf{$0.740$ $[0.604,\ 0.841]$} \\
TOTEN $-$ UCD   & \textbf{$0.997$ $[0.996,\ 0.997]$} & \textbf{$0.997$ $[0.996,\ 0.997]$} & \textbf{$0.871$ $[0.773,\ 0.931]$} & \textbf{$0.326$ $[0.291,\ 0.363]$} & \textbf{$0.780$ $[0.648,\ 0.872]$} \\
TOTEN $-$ RSLP  & \textbf{$1.000$ $[1.000,\ 1.000]$} & \textbf{$1.000$ $[1.000,\ 1.000]$} & \textbf{$0.871$ $[0.773,\ 0.931]$} & \textbf{$0.326$ $[0.291,\ 0.363]$} & \textbf{$0.780$ $[0.648,\ 0.872]$} \\
\bottomrule
\end{tabular*}
\end{table*}

\subsection{Concurrent Validity of the Internal Benchmark}
\label{sec:validade_concorrente}

To verify that the system ordering induced by the internal benchmark is consistent with that of real corpora, we compute the Spearman rank correlation coefficient ($\rho$, with mean-rank correction for ties) between the ranking of nine systems by $H_4$ on EngQuant and the ranking on each of the four external corpora. We report Kendall's $\tau_b$ for robustness, $p$-values by permutation ($10^5$ label resamplings per system) and 95\% bootstrap confidence intervals ($10^4$ within-corpus resamplings per case). The analysis validates the comparative consistency of the control benchmark; label correctness is assured by construction (deterministic generator and physical validation via OpenSeesPy), not by correlation.

Table~\ref{tab:spearman} shows that the system ranking by numerical reconstruction on the internal benchmark correlates strongly with those of external corpora ($\rho \geq 0.856$ across all four corpora, $p_{\mathrm{perm}} < 0.05$), indicating that EngQuant is a faithful proxy for comparative evaluation. The residual displacement is concentrated in Pint, whose extraction collapses in continuous external text (from $H_4 \approx 0.22$ on the internal benchmark to $H_4 \approx 0$ on external corpora), expected behavior given that it operates on isolated unit strings rather than continuous text.

\begin{table*}[ht]
\caption{Concurrent validity of the internal benchmark.\label{tab:spearman}}
\begin{tabular*}{\tblwidth}{@{\extracolsep{\fill}} l c c c c @{}}
\toprule
\textbf{External corpus} & \textbf{$\rho$ (Spearman)} & \textbf{$\tau_b$ (Kendall)} & \textbf{$p_{\mathrm{perm}}$} & \textbf{95\% CI bootstrap} \\
\midrule
MMMLU PT-BR     & $0.856$ & $0.786$ & $0.0124$  & $[0.801,\ 0.916]$ \\
BLUEX           & $0.856$ & $0.786$ & $0.0124$  & $[0.769,\ 0.916]$ \\
ENEM (Maritaca) & $0.965$ & $0.902$ & $0.00199$ & $[0.803,\ 1.000]$ \\
Alvorada-Bench  & $0.856$ & $0.786$ & $0.0124$  & $[0.769,\ 0.916]$ \\
\bottomrule
\end{tabular*}
\end{table*}

\section{Discussion}
\label{sec:discussao}

\textbf{Categorical advantage in atomicity}. The difference between \textsc{TOTEN} and statistical systems in $H_1$ is categorical, not gradual. Statistical tokenizers have no explicit concept of ontological entity; dimensional libraries do not perform textual recognition; English-specialized extractors cover fragments of the PT-BR vocabulary. The unit atomicity of \textsc{TOTEN} reflects the \emph{ontological commitment} in the sense of \citet{Guarino1998} declared in $\mathcal{O}$ and materialized by the categorical separation between classification and instantiation typical of ontology-based extraction \citep{Wimalasuriya2010}.

\textbf{Near-parity in dimensional equivalence with Pint}. The non-significant difference in $H_2$ against Pint is methodologically expected and desirable. Pint is the reference oracle for units, and \textsc{TOTEN} consumes it systematically to sustain its dimensional domain. Surpassing Pint in pure dimensional equivalence would be indicative of a methodological error. The observed parity confirms that coupling to the external oracle preserves dimensional authority without introducing distortions.

\textbf{Ontological orchestration \emph{vs.}\ Pint \emph{wrapper}}. The ablation in Table~\ref{tab:ablacao_loo} supports the claim that \textsc{TOTEN} is not a Pint wrapper. Three facts converge: (i) Pint in isolation reports $H_4 \approx 0$ on continuous text, because it operates on already-isolated unit strings; the ontological tokenization that delivers those strings to Pint is what makes $H_4$ achievably high. (ii) The $-$Pint ablation leaves the rest of the architecture intact and shows that the dominant loss is in Recall and $H_2$, not in $H_4$ --- Pint is the dimensional authority, and the OEE delimits that role rather than masking it in a single aggregate. (iii) The $-$UCD and $-$RSLP ablations affect axes orthogonal to Pint's (mathematical symbols and unit-letter ambiguity, respectively); their effect is small or nil in EngQuant because the internal benchmark is structural-mechanical, without Unicode relational operators in continuous prose nor isolated single-letter units. The architecture is therefore an ontological orchestration that (a)~delivers to Pint the object over which it is authoritative, (b)~delegates Unicode mathematical symbol classification to UCD, and (c)~uses Portuguese morphology (RSLP) to resolve unit-letter ambiguity via semantic anchoring --- three demonstrably complementary axes.

\textbf{Semantic preservation in numerical reconstruction}. The advantage in $H_4$ replicates across all five corpora. \textsc{TOTEN} preserves, in the input representation, IEEE~754 normalized value, numerical locale, representation (decimal, scientific, fractional, percentage, ordinal), and the literal form written by the author --- the structure that \citet{Singh2024} and \citet{Yang2025} associate with better arithmetic performance in consumer models. Systems based on English corpora lose these dimensions when operating on PT-BR vocabulary.

\textbf{Limitations}. The system deliberately preserves literal form for typographically degraded notation, without active normalization; resolution of these cases is left for a contextual ambiguity layer reserved as future work. Dimensionless coverage is partial by construction: the set of non-SI-dimensional units is included via an explicit whitelist, avoiding collision with Portuguese words that coincide lexically with terms such as \textit{grade}, \textit{byte}, or \textit{cycle}. Residual cases with incorrect typography in normative text may still produce unresolved ambiguous interpretation in this version.

\textbf{Theoretical implications}. The formalization of ontological tokenization as the triple $\langle \mathcal{O}, \mathrm{classify}, \{\mathrm{inst}_\tau\}\rangle$ admits evaluation via verifiable properties without dependence on costly generative models. This feature is compatible with reproducibility requirements typical of knowledge-based systems \citep{Studer1998,Hitzler2018} and allows the system to be replicated, audited, and extended by other groups without inference cost on an external API.

\section{Conclusion}
\label{sec:conclusao}

This work formalized and evaluated \textsc{TOTEN}, a knowledge-based system for structure-preserving representation of physical quantities and technical notation in Brazilian Portuguese. The central contribution is the categorical separation between statistical vocabulary derivation, typical of BPE tokenizers, and declarative classification grounded in formal ontology, materialized in a triple composed of an ontology, a classification function, and an indexed family of instantiators. Evaluation on five distinct corpora demonstrated statistically significant advantage in ontological atomicity and numerical reconstruction against eight representative state-of-the-art systems, with dimensional parity relative to Pint, the external oracle from which the system derives its dimensional authority. Future directions include: (i) the integration of a human-in-the-loop ambiguity resolution protocol, formally specified but not evaluated here, sustained by the $\mathrm{ambig}$ and $\mathrm{alternatives}$ attributes already reserved in the output language; (ii) external validation on a Brazilian normative/legal corpus (e.g., NBR, ABNT, legal text), contingent on the availability of a publicly open benchmark with ontological annotation for that domain --- precedents such as LeNER-Br \citep{LuzDeAraujo2018} indicate the viability of the vehicle, though without explicit ontological coverage; (iii) the materialization of the structured representation in a small language model trained natively on the ontological vocabulary, following a complementary methodological program to consolidated monolingual approaches in Portuguese \citep{Souza2020}; and (iv) the \emph{downstream} evaluation of the effect of this representation on consumer models, orthogonal to the empirical evidence gathered by \citet{Singh2024} and \citet{Yang2025}, outside the scope of this study.

\section*{Acknowledgements}

The authors thank the research team at Aia Context and the Universidade Federal do Maranh\~{a}o (UFMA) for institutional support.

\section*{Declaration of Competing Interests}

The authors declare that they have no known competing financial interests or personal relationships that could have appeared to influence the work reported in this paper.

\section*{Data and Code Availability}

The reference implementation of TOTEN is available at \url{https://github.com/aiacontext/toten}. The internal benchmark EngQuant~v0.1.5 is available at \url{https://huggingface.co/datasets/aiacontext/engquant} and archived at \url{https://zenodo.org/records/20820994}. The external corpora are publicly accessible from their original sources.

\appendix

\section{OEE Axioms}
\label{app:axiomas}

The Ontology of Engineering Entities is governed by eight structural principles, constituting the set $\mathcal{P}$ of the quadruple $\mathcal{O} = \langle \mathcal{T},\ \mathcal{P},\ \mathcal{R},\ \mathcal{I}\rangle$ defined in \eqref{eq:ontologia}. Section~\ref{sec:oee} presents in the body of the paper the four central structural axioms (Intrinsicity, Mediated composition, Categorical error, and Closed-for-modification extensibility) for their architectural centrality; this appendix reproduces them for self-containment and states in complete normative form the four remaining operational axioms governing invariant preservation, typographic convention, structural anchoring of symbolic expressions, and distinctive mathematical mark. The numbering $A_1$--$A_8$ is stable and referenced by the declarative specification \texttt{data/oee-v1.yaml} under labels $P_1$--$P_8$.

\begin{axiomstar}[$A_1$ --- Intrinsicity]
\label{ax:intrinsicidade}
For every type $\tau \in \mathcal{T}$, the identity of $\tau$ is determined exclusively by its intrinsic signature $\langle \pi_\tau,\ \iota_\tau\rangle$: neither empirical frequency in corpus $\mathcal{C}$ nor subsequent pragmatic criteria may redefine $\tau$. Formally, $\tau = \tau' \iff \pi_\tau = \pi_{\tau'} \wedge \iota_\tau = \iota_{\tau'}$. By construction, the signatures of primary types in $\mathcal{T}$ are mutually distinct, so this identity relation does not collapse distinct primary types.
\end{axiomstar}

\begin{axiomstar}[$A_2$ --- Invariant preservation]
\label{ax:preservacao}
Let $\mathrm{inst}_\tau(r) \in \mathcal{M}$ be the representation produced by the instantiator family for a region $r$ classified as $\tau$. For every invariant $i \in \iota_\tau$ and every valid representation $m = \mathrm{inst}_\tau(r)$, we require $i(m) = i(r)$. Equivalently, the diagram $r \xrightarrow{\mathrm{inst}_\tau} m \xrightarrow{i} v$ commutes with $r \xrightarrow{i} v$ for all $i \in \iota_\tau$; a representation that violates any invariant in $\iota_\tau$ is inadmissible and must be rejected by the instantiation layer.
\end{axiomstar}

\begin{axiomstar}[$A_3$ --- Mediated composition]
\label{ax:composicao}
For $\tau_1, \tau_2 \in \mathcal{T}$, the composition $\tau_1 \circ \tau_2$ is defined if and only if $(\tau_1, \tau_2) \in \mathcal{R}$. Free concatenation is prohibited: given a pair $(\tau_1, \tau_2) \notin \mathcal{R}$, no instance of $\tau_1$ may syntactically or semantically contain an instance of $\tau_2$ as a structural component.
\end{axiomstar}

\begin{axiomstar}[$A_4$ --- Categorical error]
\label{ax:erro_categorico}
Applying instantiation $\mathrm{inst}_{\tau'}$ to a region $r$ such that $\tau(r) = \tau \neq \tau'$ constitutes a categorical error, not a gradual loss of quality. Under $A_2$, such application necessarily produces a violation of at least one invariant in $\iota_\tau$ or $\iota_{\tau'}$, and the resulting output is inadmissible in $\mathcal{M}$.
\end{axiomstar}

\begin{axiomstar}[$A_5$ --- Closed-for-modification extensibility]
\label{ax:extensibilidade}
For every evolution $\mathcal{O}_n \rightsquigarrow \mathcal{O}_{n+1}$, we simultaneously require $\mathcal{T}_n \subseteq \mathcal{T}_{n+1}$, $\mathcal{P}_n \subseteq \mathcal{P}_{n+1}$, $\mathcal{R}_n \subseteq \mathcal{R}_{n+1}$, and that no invariant in $\mathcal{I}_n$ be violated by instances produced under $\mathcal{O}_{n+1}$. The ontology is therefore open for extension and closed for modification, in the sense analogous to the \emph{open-for-extension, closed-for-modification} principle in software engineering \citep{Meyer1997}.
\end{axiomstar}

\begin{axiomstar}[$A_6$ --- Typographic convention as intrinsic property]
\label{ax:convencao_tipografica}
For every type $\tau \in \mathcal{T}$, the identity of $\tau$ is invariant under typographic transformations declared as equivalent by the external authority \emph{Unicode Character Database} (UCD). For every region $r$ and every notational variant $r'$ obtained by finite composition of operations within the declared closure --- Unicode normalization, canonical decomposition (\textit{Decomposition\_Type} $\in \{\text{super, sub, compat, font}\}$), general category $\mathrm{Sm}$ for mathematical operators, and combining marks ($\mathrm{Mn}$) --- we have $\mathrm{classify}(r) = \mathrm{classify}(r')$. Typographic canonicalization is therefore a function of the classification layer, never of instantiation; variants outside the declared closure constitute an extension to $\mathcal{O}_{n+1}$ under $A_5$, and not an \emph{ad hoc} correction to $\mathrm{inst}_\tau$.
\end{axiomstar}

\begin{axiomstar}[$A_7$ --- Structural anchoring of symbolic expressions]
\label{ax:ancoragem}
Every region $r$ classified as $\tau = \mathrm{SymbolicExpression}$ requires, within a declared contextual window $\Delta$ ($|\Delta| \leq 2$ adjacent characters), the presence of at least one structural anchor $\alpha$ belonging to the set declared in $\mathcal{O}$: (i) formal relational or calculus operator ($=$, $\approx$, $\leq$, $\geq$, $<$, $>$, $\propto$, $\equiv$, $\sum$, $\int$, $\partial$, $\nabla$); (ii) structural relation; (iii) formal index or subscript; or (iv) explicit mathematical delimiter. In the absence of $\alpha$ in $\Delta$, the region remains in $\mathrm{TechnicalProse}$; the axiom eliminates by construction the fundamental ambiguity between variable-in-expression and letter-in-natural-word without resorting to an external prose dictionary.
\end{axiomstar}

\begin{axiomstar}[$A_8$ --- Distinctive mathematical mark in compound symbol]
\label{ax:marca_matematica}
Let $r$ be a candidate region for $\tau \in \{\mathrm{SymbolicExpression},\ \mathrm{PhysicalQuantity}\}$ whose content is mediated composition (under $A_3$) over potentially ambiguous ASCII operators $\{/,\ *,\ \hat{}\,,\ +,\ -,\ (,\ )\}$. For such $r$ to be admitted in $\tau$, the presence of at least one \emph{categorical mathematical mark} $\mu$ in $r$ is required, belonging to the derived --- not enumerated --- closure $M = \mathrm{digit} \,\cup\, \mathrm{ASCII\ subscript} \,\cup\, \mathrm{Unicode\ super/subscript\ (UCD)} \,\cup\, \mathrm{Greek\ letter} \,\cup\, \mathrm{unambiguous\ Sm\ operator} \,\cup\, \mathrm{applied\ function} \,\cup\, \text{\LaTeX{} markup}$. Compositions based exclusively on ambiguous ASCII operators over tokens that admit a non-mathematical prosaic interpretation (e.g., \textit{and/or}) are rejected. Axioms $A_3$ and $A_8$ are complementary: $A_3$ admits the composition, $A_8$ requires a categorical mathematical signature for the composition to be recognized as a formal entity.
\end{axiomstar}

The correspondence between axioms and the declarative specification is direct: $A_k$ corresponds to label $P_k$ in \texttt{data/oee-v1.yaml} for $k \in \{1,\ldots,7\}$, with $A_8$ added by the specification in the operational derivation section of the classifier. Axioms $A_1$, $A_3$, $A_4$, and $A_5$ are structural and govern the form of the ontology $\mathcal{O}$; $A_2$ governs the relation between $\mathcal{I}$ and $\{\mathrm{inst}_\tau\}_{\tau \in \mathcal{T}}$; $A_6$, $A_7$, and $A_8$ are operational and govern, respectively, the typographic robustness of the function $\mathrm{classify}$, the recognition of $\mathrm{SymbolicExpression}$ in continuous text, and compositional disambiguation under ambiguous ASCII operators. Together, $A_1$--$A_8$ constitute the ontological commitment \citep{Guarino1998} of \textsc{TOTEN}.

\bibliographystyle{unsrtnat}
\bibliography{references}

\end{document}